%% file: neurips_2020.tex
\documentclass{article}

\usepackage{wrapfig}

\usepackage[nonatbib, final]{neurips_2020}

\usepackage[utf8]{inputenc} 
\usepackage[T1]{fontenc}    
\usepackage{hyperref}       
\usepackage{url}            
\usepackage{booktabs}       
\usepackage{amsfonts}       
\usepackage{nicefrac}       
\usepackage{microtype}      

\usepackage[ruled,vlined]{algorithm2e}
\usepackage{booktabs}
\usepackage{amssymb}
\usepackage{graphicx,subfigure}
\usepackage{amsmath}
\usepackage{xcolor}

\input{math_config}

\input{other_configs}

\title{\PaperNickname: \PaperDescription}
\author{%
  Tatiana L\'{o}pez-Guevara$^{1,2}\;$
  Michael Burke$^{2}\;$
  Nicholas K. Taylor$^{1}\;$
  Kartic Subr$^{2}$
  \\
  $^{1}$School of Mathematics and Computer Science, Heriot-Watt University\\
  $^{2}$School of Informatics, University of Edinburgh\\
  \texttt{t.l.guevara@ed.ac.uk} \\
}

\begin{document}

\maketitle


\begin{abstract}
  Model-free reinforcement learning (RL) is a powerful tool to learn a broad range of robot skills and policies. However, a lack of policy interpretability can inhibit their successful deployment in downstream applications, particularly when differences in environmental conditions may result in unpredictable behaviour or generalisation failures. 
  As a result, there has been a growing emphasis in machine learning around the inclusion of stronger inductive biases in models to improve generalisation.
  This paper proposes an alternative strategy, \MakeLowercase{\PaperDescription} 
  (\PaperNickname), which seeks to identify the inductive biases or idealised conditions of operation already held by pre-trained policies, and then use this information to guide their deployment. 
  \PaperNickname\ uses Masked Autoregressive Flows to fit distributions over the set of conditions or environmental parameters in which a policy is likely to be effective. This distribution can then be used as a policy certificate in downstream applications. We illustrate the use of \PaperNickname\ across a two environments, and show that substantial performance gains can be obtained when policy selection incorporates knowledge of the inductive biases that these policies hold.

\end{abstract}


\section{Introduction}

The proliferous development of methods that use reinforcement learning~\cite{andrychowicz2020learning, Chebotar2018ClosingTS, Bagaria2020OptionDU, sutton1999between} or other strategies~\cite{wang2018active, wang2020learning, guevara2017adaptable, tostir} for learning robotic manipulation skills has enabled the creation of exciting  libraries of potentially viable policies to solve  many common tasks. 
However, an extra layer of interpretability to identify when such policies are likely to succeed is important for their successful deployment, where uncertainty around environmental conditions and parameters may exist. This has motivated the inclusion of strong inductive biases \cite{LeGuen,Li2020Learning,jaques2020newtonianvae} or alternative forms of certification over these models~\cite{wang2018active}.

\begin{figure}[h!]
    \centering
    \includegraphics[width=0.9\linewidth]{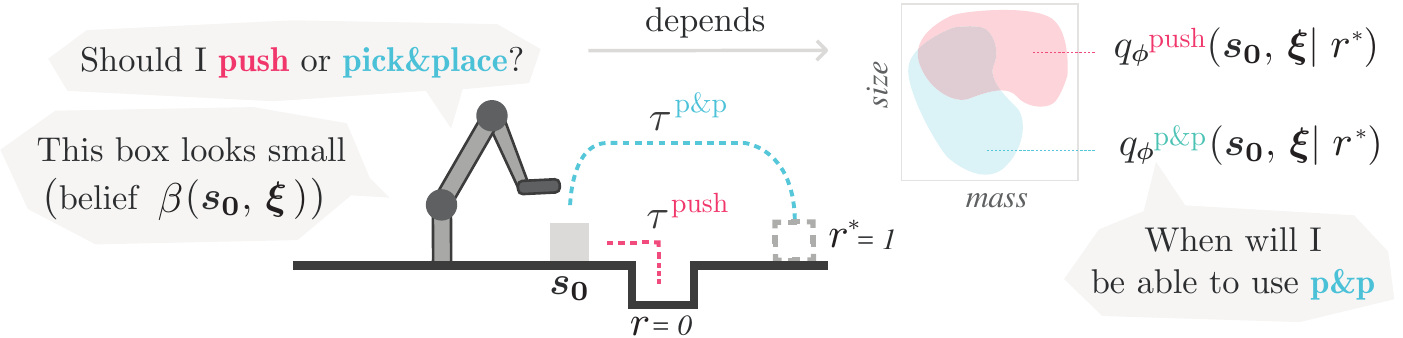}
    \caption{\label{fig:example} 
    We can use the set of initial conditions and parameters learned using \PaperNickname\ to select how to move an object based on the agent's belief. It is better to pick and place small objects that start at the left of the hole and push bigger or heavier objects. }
    \vspace{-4mm}
\end{figure}

For example, consider an agent that wants to move an object as shown in~Figure-\ref{fig:example}. The agent can either push or pick and place, but choosing the appropriate strategy depends on the belief around the object being manipulated:
i.e pick and place is preferable for lighter objects whereas push is better for bigger or heavier ones.
We propose a methodology to enable interpretability of such expert systems that choose from 
different pre-trained policies.
Specifically, we focus on how to infer the posterior probability ($\paramdistC$) of the model parameters and initial conditions ($\stateoparam$) given evidence of the rewards ($\reward$) from an arbitrarily chosen policy.  These posteriors 
($\paramdistC^{\mathrm{push}}$, $\paramdistC^{\mathrm{p\&p}}$, etc.)
may be used to reason about  viability and preferability or to counterfactually query environment settings and conditions in which a policy may fail. 
We perform \MakeLowercase{\PaperDescription} (\PaperNickname ), to identify inductive biases or idealised conditions of operation \textit{already held by pre-trained policies}, and then use this information to guide their deployment. 

\section{Related Work}
{\PaperNickname} fits posterior distributions over environment states or settings, conditioned on rewards. This is similar to recent work that attempts to learn models of preconditions for planning \cite{wang2018active, wang2020learning}, by generating samples of valid initial states for a given policy based on the context of a scene. Here, samples are drawn from Gaussian Processes using level sets, so this approach is only applicable to low-data regimes.
The use of interpretable policy posteriors for task selection is also related to that of initiation sets within the Options Framework
~\cite{sutton1999between, konidaris2009skill, Bagaria2020OptionDU} or abstractions in skill segmentation~\cite{niekum2015learning, konidaris2011cst}. However, these sets are usually discovered by learning classifiers that identify, from which observable states, an option was successfully triggered, without any notion of uncertainty. In contrast, by fitting a full posterior over environments, {\PaperNickname} also allows for counterfactual reasoning of the form, 'Where won't this policy work?'.

The Sim2Real (S2R) community~\cite{weng2019DR, andrychowicz2020learning, akkaya2019solving, Ramos2019BayesSimAD, Chebotar2018ClosingTS} has developed a number of methods to iteratively learn a policy and fit distributions over environment parameters that can be used to model a particular instance of a real world, often relying on Gaussianity~\cite{Ramos2019BayesSimAD, Chebotar2018ClosingTS} or independence assumptions~\cite{Ramos2019BayesSimAD, andrychowicz2020learning} regarding the underlying distribution.
Of particular interest here is 
Likelihood Free Inference (LFI),
which is often used for simulation alignment to infer which simulation parameter generated a particular real trajectory or observation
~\cite{cranmer2019frontier, durkan2020contrastive, Ramos2019BayesSimAD, Gutmann2016a, papamakarios_epsilon_free, papamakarios2018sequential, Greenberg2019AutomaticPT}.
Recent advances in LFI have shown exciting results in the quality of the estimated posterior by relying on contrastive losses and sequential estimation~\cite{durkan2020contrastive, Greenberg2019AutomaticPT}.
Our work also relies on sequential LFI, but focuses on estimating a full posterior distribution over initial conditions and environment parameters that might be partially observable during deployment. Importantly, by doing so we allow for counterfactual reasoning and the use of a broad range of easily computed metrics that incorporate uncertainty, like the proposed cross entropy criterion for task selection in expert systems. Existing sample-based approaches producing an intractable posterior do not directly allow this. By framing our problem within the LFI framework, we take advantage of recent algorithmic developments in inference and density estimation.


\section{\PaperNickname}
{\PaperNickname} is a sequential a method to learn interpretable posteriors over the set of initial conditions or environmental parameters in which a policy is likely to be effective.


\subsection{Problem formulation}

Let $\MDP{}$ define a 
Markov Decision Process,
composed of states $\state \in \StateS$, actions $\action \in \ActionS$ and  a reward map $\RewardS : \StateS \to \R$. The distribution of possible starting states is defined by $\stateosim$. Transitions are determined by $\statettsima$, where $\param~\sim~\paramprior{}$ represent the parameters that affect the dynamics of the environment (extrinsic/intrinsic parameters of the objects involved in the task). Note that $\paramprior{}$ implicitly defines a \textit{class of objects}  from which particular object instances can be sampled (i.e, objects with different shapes, masses, etc.). 
An agent can act on such environments according to a policy distribution $\actionsim$ parametrized by $\policyparC$.
Given a initial state $\stateo$, parameter $\param$ and a policy $\policyparC$, the agent will observe a trajectory after each interaction as $\trajdist = \prod_{t=0}^{T} \statettdist \stateodist$.
The observed rewards will thus be distributed according to
$\reward \sim \trajdistF{\RewardS}$.
Let $\paramdistC(\stateoparam|\reward)$ represent the variational approximator of the distribution over initial states and parameters conditioned on the reward, i.e an autorregressive flow parametrised by $\paramdistparC$. 
\begin{wrapfigure}{r}{0.65\textwidth}
\vspace{-1mm}
    \centering
    \includegraphics[width=0.65\textwidth]{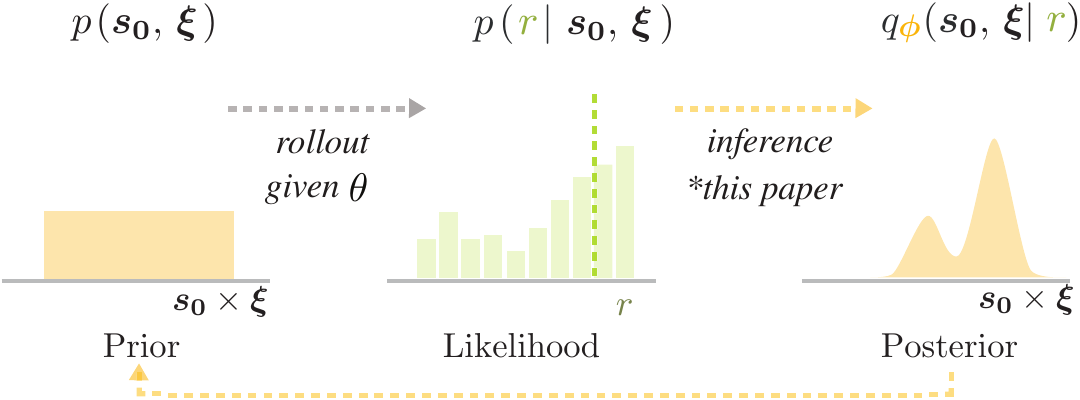}
    \caption{\label{fig:overview} Sequential Estimation of the Conditional Posterior on Environment Parameters.  }
    \vspace{-4mm}
\end{wrapfigure}
Let $\beliefdist$ represent the belief distribution over initial states and parameters of the object the agent wants to manipulate.

\label{sec:form:posterior}

Given a policy, we wish to learn under which initial conditions and object instances the policy is most likely to succeed. Essentially, we are interested in the posterior distribution over instances of the environment that will yield high reward.





\begin{equation}
\label{eq:rel:inference_via_reward}
    \paramdistparC^*(\policyparC) = 
        \argmin_{\paramdistparC} 
        \E_{
            \substack{
                \stateoparam \sim \stateoparamdist\\
                \reward \sim \rewarddist
                }
            } 
        \bigg[ \;
            -\log \paramdistC(\stateoparam|\reward) \; 
        \bigg] \;
\end{equation}

We sequentially narrow down the posterior across $\iterlims$ iterations and update the prior of the next iteration with the conditioned posterior learned in the previous (Algorithm~\ref{alg:sequential} / Figure~\ref{fig:overview}). At each iteration, we collect a dataset $\dseti{\iter}$  from $\rollout$ observed rollouts
$\{ (\stateoparam, \reward)^{(\rollout)} \}_{\rollout=1}^{\nrollouts}$ and optimize Eq~\ref{eq:rel:inference_via_reward} to fit the variational approximator of the posterior $\paramdistC$.  We used the Automatic Posterior Transformation~\cite{Greenberg2019AutomaticPT} algorithm to optimize Eq~\ref{eq:rel:inference_via_reward}. 
The form of $\paramdistC$ should be flexible enough to capture complex correlations across parameters (i.e. Figure~\ref{fig:rel:fig:posterior}-\textit{Right}). Therefore assuming a diagonal covariance~\cite{Ramos2019BayesSimAD} might not be enough. 
Masked Autoregressive Flow (MAF)~\cite{papamakarios2017masked} is a powerful density estimator method that excels at modelling complex distributions, compared to other neural estimators, for inference~\cite{papamakarios2017masked, Greenberg2019AutomaticPT}. 
We model our posterior distribution $\paramdistC$ using MAF.

\vspace{-2mm}
\subsection{Posterior for Task Selection}

The learned posterior distribution $\paramdistC$ can also be used as a policy certificate in downstream applications like task selection. 
Consider the case where the agent wants to manipulate an instance of an object for which it has an initial belief $\beliefdist$ (Figure~\ref{fig:example}). Assume it already knows how to perform a task in different ways $\policyparC^{(\taskid)}$ and that we learned their respective  posteriors $\paramdistparC^{(\taskid)}$ as described in Section \ref{sec:form:posterior}.




We can use $\paramdistparC$ to select which manipulation skill to use by computing the cross-entropy with respect to the prior belief of the object:

\begin{equation}
    \begin{aligned}
        \taskid^* &= \argmax_{\taskid} \; 
                        \E_{
                        \stateoparam \sim \beliefdist
                        } 
                        \bigg[
                            \log \paramdistC^{\taskid}(\stateoparam|\reward^{*}) 
                        \bigg]
    \end{aligned}
\end{equation}

\vspace{-2mm}
\section{Experiments and Results}
Our experiments aim to answer the following questions: (1) What is the quality of the learned posterior in terms of interpretability? 
(2) How useful is the posterior over the environment parameters for task selection? 

\begin{figure}[h!]
    \vspace{-3mm}
    \centering
    \includegraphics[width=0.9\linewidth]{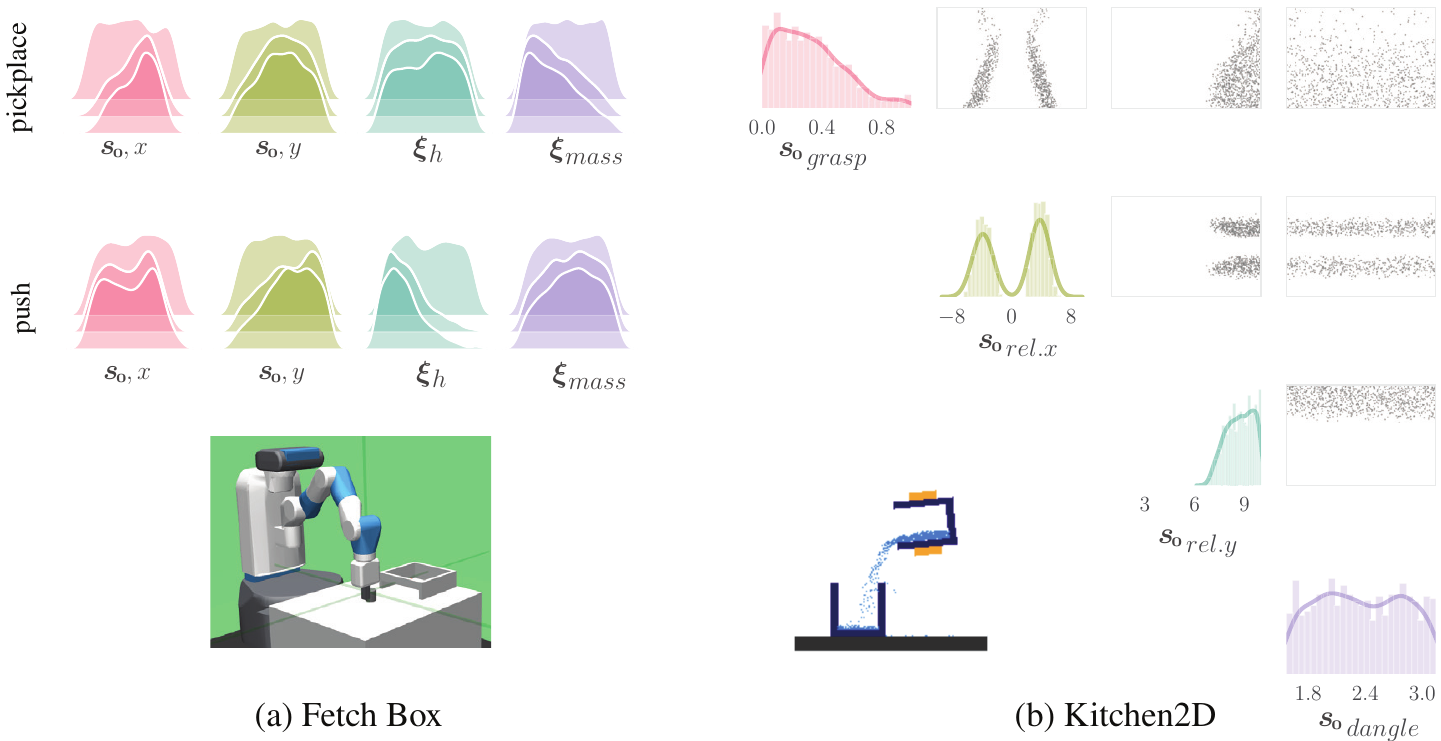} 
    \caption{\label{fig:rel:fig:posterior} (Left) Evolution of the marginals of the posterior at iterations $\iter=1,7,15$ for pickplace and push in the Fetch Box environment. (Right) Samples from the posterior at the last iteration $\iter=10$  in the Kitchen2D environment, visualized as pairwise relations between variables.} \vspace{-2mm}
\end{figure}

We tested our method on two different environments: Kitchen2D-Pouring~\cite{wang2018active} with their predefined dense reward of $exp(2 * ( x * 10 - 9.5)) - 1$, where $x$ is the proportion of liquid particles that fall outside the target container. Here $\param=\emptyset, \stateo=\{ grasp, rel.x, rel.y, dangle \}, \nrollouts=300$ and $\niters=10$. To evaluate the task selection downstream task, we used pre-trained policies for {\verb pickplace } and {\verb push } using Hindsight Experience Replay (HER)~\cite{andrychowicz2017hindsight}\footnote{ \url{https://github.com/TianhongDai/hindsight-experience-replay}}. We created a new environment in OpenAI~\cite{brockman2016openai}, FetchBox,
where the goal is to move a puck inside a box with a hole at the bottom (Figure~\ref{fig:rel:fig:posterior}-\textit{Left, Bottom}). The reward is $1$ if the puck is inside the box at the end of the episode and $0$ otherwise. We chose $\stateo = \{ x, y\}$ to be the initial position of the puck in the table, and the parameters of the puck to be a combination of mass, width, height and friction (Table \ref{tb:rel:params:fetch} shows the used ranges) $\param \subset \{mass, h, w, fr\}$, $\nrollouts=500$ and $\niters=15$. 


Figure~\ref{fig:rel:fig:posterior}-\textit{Left} shows evolution of the marginals of the posterior at iterations $\iter=1,7,15$ for $\paramdistparC^{pickplace}$ and $\paramdistparC^{push}$ using 5000 samples. We can observe 
that, even after 7 iterations, the posterior varies significantly between both policies, especially with respect to $\param_{mass}$ and $\param_{h}$. This indicates that $\policyparC^{pickplace}$ works best for pucks with mid to low mass and $\policyparC^{push}$ with short pucks due to the obstacle in the box.
We also present 1000 samples of the final posterior distribution for the Kitchen2D environment in Figure~\ref{fig:rel:fig:posterior}-\textit{Right}. We can observe a multimodal correlation between the relative positions of the containers and where to grasp the source container, as well as the irrelevance of the final angle of the cup for this task.

\begin{figure}[h!]
    \centering
    \vspace{-3mm}
        \includegraphics[height=10em]{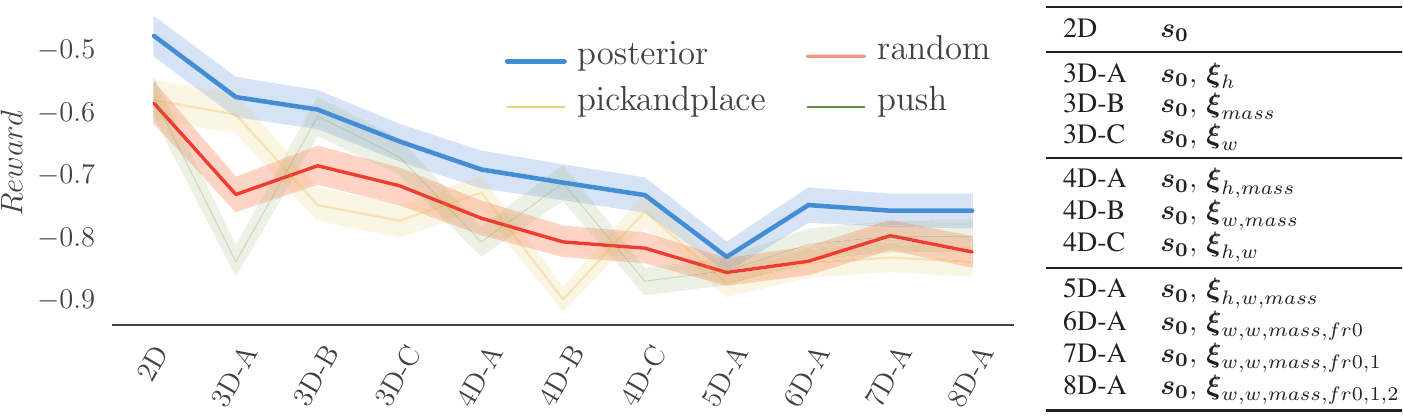} 
    \caption{\label{fig:rel:mtask:d3} Average rewards for different task selection methods over the initial position of the puck on the table $\stateo$ and different subsets of parameters $\param$ for 1k randomly initialized object beliefs.} \vspace{-4mm}
\end{figure}

To test the usefulness of the posterior for task selection, we used the FetchBox environment to move a puck into a  box~(Figure~\ref{fig:rel:fig:posterior}-\textit{Left}).  
The goal is to select between $\policyparC^{push}$ and $\policyparC^{pickplace}$ policies to move a puck for which the agent has a belief $\beliefdist$. We sampled 1000 different beliefs of the puck according to $\beliefdist = \mathcal{N}( \; \param \; | \;\mub, \; \Sigmab \;)$ with $\mu^{(j)} = 0.5 $ and $\Sigmab^{(j,j)} = 0.7$ and off-diagonal $0$. The belief is then rescaled to valid ranges according to Table \ref{tb:rel:params:fetch}. We use random selection as the baseline and also show always pick or always push as guidelines. We show different combinations of parameters with different dimensionalities on the x-axis and the average reward on the y-axis. All configurations of methods and parameters were executed 1000 times. 
Figure~\ref{fig:rel:mtask:d3} shows the performance gained by using the learned posterior for task selection.

\section{Conclusions}

We presented \PaperNickname, a method to learn the inductive biases held by a policy, by fitting a distribution over the set of conditions or environmental parameters in which a policy is likely to be effective. We showed that this posterior provides an
additional layer of interpretability to preexisting policies, allows counterfactual reasoning, and facilitates task selection in expert systems without requiring an intractable posterior to be computed using an expectation over samples. In future work we  plan to apply {\PaperNickname} to curriculum learning and domain adapation.


\newpage
\small
\bibliographystyle{plain}
\bibliography{neurips_2020}  

\section*{Appendix}
\subsection{\PaperNickname Algorithm} 
\begin{algorithm}[H]  
\DontPrintSemicolon
  
    \KwIn{$\stateoparamdist, \trajdist$}  
    \KwOut{$\paramdistparC^*$}

    \For{$\iterlims$}{
        \For{$\rolloutlims$}{
        \nl Sample $\stateoparam \sim \stateoparamdist$ \;
        \nl Sample $\trajsim$ \;
        \nl Compute $\reward =  \RewardS(\Tp{})$\;
        \nl Add to dataset $\dseti{\iter} = \dseti{\iter-1} \cup \{\;
            (\stateoparam, \reward) \; \}$ \;
        }
        \nl Optimize $\mathcal{L}(\paramdistparC(\policyparC))$  using $\dseti{\iter}$ (Eq~\ref{eq:rel:inference_via_reward}) \;
        \nl Get $\reward^{*}$ from $\dseti{\iter}$ \; 
        \nl Set $\stateoparamdist$ = $\paramdistC(\stateoparam | \reward^{*})$
    }
\caption{\label{alg:sequential}\bf \PaperNickname: Sequential Refinement of the Posterior}
\end{algorithm}

\subsection{Range of Parameter Priors used in the Experiments}

In all experiments we used a uniform prior for $\stateoparamdist$ with the following parameter limits:

\begin{table}[h]
    \begin{center}
        \begin{footnotesize}
            \begin{tabular}{l l c c c c c}
                \toprule
                & Description & Notation & Name in Sim. & \multicolumn{2}{c}{Range} \\
                \cline{5-6}
                &             &          &              & $\param_{min}$ & $\param^{*}$ \\
                \toprule
                Kitchen2D${}^{*}$ & & & & & \\ 
                & Grasp ratio & $\stateo_{grasp}$ & {\verb grasp_ratio } & 0.0   & 1.0 \\
                & Relative x & $\stateo_{rel.x}$ & {\verb rel_x } & -10.0   & 10.0 \\
                & Relative y & $\stateo_{rel.y}$ & {\verb rel_y } & 1.0   & 10.0 \\
                & Final angle & $\stateo_{dangle}$ & {\verb dangle } & $0.5*\pi$   & $\pi$ \\
                \midrule
                Fetch OpenAI${}^{+}$ & & & & & \\
                & Mass                & $\param_{mass}$ & {\verb body_mass }     & 1.0    & 20.0  \\
                & Size x              & $\param_{w}$ & {\verb geom_size[0] }     & 0.02   & 0.045 \\
                & Size y              & $\param_{h}$ & {\verb geom_size[1] }     & 0.02   & 0.03  \\
                & Tangential Friction & $\param_{fr0}$ & {\verb geom_friction[0] } & 0.1   & 1.0  \\
                & Torsional Friction  & $\param_{fr1}$ & {\verb geom_friction[1] } & 0.1   & 1.0  \\
                & Rolling Friction    & $\param_{fr2}$ & {\verb geom_friction[2] } & 0.1   & 1.0  \\
                \bottomrule
            \end{tabular}
            \caption{Object Parameter Priors used in ${}^{*}$Kitchen2D~\cite{wang2018active}  (Box2D Engine) and ${}^{+}$Fetch (Mujoco Engine) environments (Name in Sim corresponds to the name of the parameter in each engine/wrapper. }
            \label{tb:rel:params:fetch}
        \end{footnotesize}
    \end{center}
\end{table}

\end{document}

%% file: math_config.tex
\DeclareMathOperator*{\argmin}{argmin}
\DeclareMathOperator*{\argmax}{argmax}
\DeclareMathOperator*{\E}{\mathop{\mathbb{E}}}

\definecolor{zcyan}{RGB}{0,187,214} 
\definecolor{zpink}{RGB}{225,0,124} 
\definecolor{zyellow}{RGB}{239,164,0} 
\definecolor{zgreen}{RGB}{34,139,34} 
\definecolor{zgray}{RGB}{70,70,70} 

\newcommand{\mub}  {{\ensuremath{ {\boldsymbol{\mu}} }}}
\newcommand{\Sigmab}  {{\ensuremath{ {\boldsymbol{\Sigma}} }}}

\newcommand{\distsymb}  {{\ensuremath{ {p} }}}

\newcommand{\param}  {{\ensuremath{ {\boldsymbol{\xi}} }}}

\newcommand{\paramdistpar}  {\ensuremath{ {\boldsymbol{\phi}} }}

\newcommand{\paramprior}[1]  {\ensuremath{ \distsymb^{#1}(\param) }}

\newcommand{\paramdistparC}  {\textcolor{zyellow}{\paramdistpar}}

\newcommand{\paramdistC}  {\ensuremath{ q_{\paramdistparC} }}

\newcommand{\Tp} [1]{\ensuremath{\tau_{#1}}}

\newcommand{\state}  {\ensuremath{ \boldsymbol{s} }}
\newcommand{\stateo}  {\ensuremath{ \boldsymbol{s_0} }}
\newcommand{\statet} {\ensuremath{ \boldsymbol{s}_t  }}
\newcommand{\statett} {\ensuremath{ \boldsymbol{s}_{t+1}  }}

\newcommand{\R}{\ensuremath{\mathbb{R}}}
\newcommand{\MDP}[1] {\ensuremath{\mathcal{M}^{#1}}}
\newcommand{\StateS}  {\ensuremath{ \mathcal{S} }}
\newcommand{\ActionS}  {\ensuremath{ \mathcal{A} }}

\newcommand{\RewardS} {R}
\newcommand{\statettdist}  {\ensuremath{ \distsymb(\statett|\statet, \policydista; \param) }}
\newcommand{\statettdista}  {\ensuremath{ \distsymb(\statett|\statet, \action; \param) }}
\newcommand{\statettsima}  {\ensuremath{ \statett~\sim~\statettdista }}

\newcommand{\stateodist}  {\ensuremath{ \distsymb(\stateo) }}

\newcommand{\stateosim}  {\ensuremath{ \stateo~\sim~\stateodist }}
\newcommand{\beliefdist} {\ensuremath{ \beta(\stateoparam) }}
\newcommand{\stateoparam} {\ensuremath{\stateo, \, \param}}
\newcommand{\stateoparamdist} {\ensuremath{ p(\stateoparam) }}

\newcommand{\taskid} {\ensuremath{i}}

\newcommand{\iter} {\ensuremath{k}}
\newcommand{\niters} {\ensuremath{N}}
\newcommand{\iterlims} {\ensuremath{\iter=1:\niters}}
\newcommand{\rollout} {\ensuremath{b}}
\newcommand{\nrollouts} {\ensuremath{B}}
\newcommand{\rolloutlims} {\ensuremath{\rollout=1:\nrollouts}}
\newcommand{\dset} {\ensuremath{\mathcal{D}}}
\newcommand{\dseti}[1] {\ensuremath{ \dset^{(#1)}  }}

\newcommand{\policypar}  {{\ensuremath{ {\boldsymbol{\theta}} }}}
\newcommand{\policyparC}  {\textcolor{black}{\policypar}}

\newcommand{\policy} {\ensuremath{\pi}}
\newcommand{\policydista} {\policy_{\policyparC}(\actiont|\statet)}

\newcommand{\trajdist}  { p(\Tp{} | \stateoparam, \policyparC) }
\newcommand{\trajdistF}[1]  { p(#1({\Tp{}}) | \stateoparam, \policyparC) }
\newcommand{\trajsim}  { \Tp{}~\sim~\trajdist }

\newcommand{\action}  {\ensuremath{ \boldsymbol{a} }}
\newcommand{\actiont} {\ensuremath{ \action{_t} }}
\newcommand{\actionsim}  {\ensuremath{ \actiont~\sim~\policydista }}

\newcommand{\reward} {\textcolor{zgreen}{\ensuremath{r}}}

\newcommand{\rewarddist} {\ensuremath{p(r|\stateoparam, \policyparC)}}



%% file: other_configs.tex
\newcommand{\PaperNickname} {IV-Posterior}
\newcommand{\PaperDescription} {Inverse Value Estimation for 
Interpretable Policy Certificates}

